\documentclass[letterpaper,10pt,journal,twoside]{IEEEtran}

\IEEEoverridecommandlockouts

\usepackage{graphicx}
\usepackage{newtxtext,newtxmath}
\usepackage{amsmath}
\usepackage{booktabs}
\usepackage{multirow}
\usepackage[caption=false,font=footnotesize]{subfig}
\usepackage[hidelinks]{hyperref}

\newcommand{\best}[1]{\textbf{#1}}
\newcommand{\second}[1]{\underline{#1}}
\newcommand{\resultci}[2]{\shortstack{#1\\{\tiny [#2]}}}

\title{Auditing Instruction--Trajectory Mismatches in Multimodal Robot Demonstrations}

\author{Simon Holk$^{1,2}$,
Ryosuke Takanami$^{1,2}$,
Tatsuya Matsushima$^{1,2}$,
Yusuke Iwasawa$^{2}$,
Yutaka Matsuo$^{2}$,
Yueh-Hua Wu$^{1}$,
and Kei Ota$^{1}$%
\thanks{Manuscript received: May 24, 2026;
Accepted: July 27, 2026.}%
\thanks{This paper was recommended for publication by Editor
Wei Pan upon evaluation of the Associate Editor and Reviewers'
comments.}%
\thanks{$^{1}$Simon Holk, Ryosuke Takanami, Tatsuya Matsushima,
Yueh-Hua Wu, and Kei Ota are with the AI Robot Association (AIRoA),
Tokyo, Japan.}%
\thanks{$^{2}$Simon Holk, Ryosuke Takanami, Tatsuya Matsushima,
Yusuke Iwasawa, and Yutaka Matsuo are with the Matsuo-Iwasawa
Laboratory, Graduate School of Engineering, The University of Tokyo,
Tokyo, Japan.
{\tt\footnotesize simon.holk@weblab.t.u-tokyo.ac.jp}}%
\thanks{\textcopyright~2026 IEEE. Personal use of this material is permitted. Permission from IEEE must be obtained for all other uses, in any current or future media, including reprinting/republishing this material for advertising or promotional purposes, creating new collective works, for resale or redistribution to servers or lists, or reuse of any copyrighted component of this work in other works.}%
}

\hypersetup{
  pdfauthor={Simon Holk; Ryosuke Takanami; Tatsuya Matsushima; Yusuke Iwasawa; Yutaka Matsuo; Yueh-Hua Wu; Kei Ota},
  pdftitle={Auditing Instruction--Trajectory Mismatches in Multimodal Robot Demonstrations},
  pdfsubject={Offline auditing of multimodal robot demonstration datasets},
  pdfkeywords={Data Sets for Robot Learning, Learning from Demonstration}
}

\begin{document}

\maketitle

\begin{abstract}
Robot demonstration datasets used to train vision-language-action policies can contain a subtle but harmful failure mode: trajectories that are behaviorally correct but paired with the wrong language instruction. We study post-hoc auditing of these \emph{Instruction--Trajectory Mismatches} (ITMs). Unlike failed rollouts, ITMs often look plausible, and can corrupt the language--behavior mapping learned by the policy. We propose \emph{Multimodal Probabilistic Fusion} (MMPF), a training-free auditing framework that treats each modality as an expert, estimates a task-label distribution from local neighborhood agreement and global prototype similarity, and then fuses modalities with predictive-entropy weighting in a product of experts. Across LIBERO benchmarks with injected instruction mismatches and noisy real-robot data, MMPF achieves the strongest overall ITM detection and label correction accuracy. We also show that auditing improves most downstream policy learning in settings where language is needed to disambiguate the task. We demonstrate in real robot experiments that our method can achieve improved policy performance and show the trade-off of filtering demonstrations compared to relabeling.
\end{abstract}


\section{INTRODUCTION}

\IEEEPARstart{R}{obot} learning is rapidly shifting from training task-specific policies on small, curated datasets to training generalist policies on large, heterogeneous collections of demonstrations \cite{brohan2022rt, zitkovich2023rt, o2024open, team2024octo, kim2024openvla, black2026pi0visionlanguageactionflowmodel, intelligence2025pi_}. As these datasets scale across tasks, labs, and embodiments, label quality can become a bottleneck. Manual verification of labels does not scale, and even a small rate of systematic annotation errors can corrupt the supervision used to train vision-language-action (VLA) models.

Most robotics curation work focuses on execution-level problems such as failed rollouts, low-quality actions, or redundant and uninformative segments \cite{agia2025cupid, hejna2025robot, dass2025datamil, chenarss25, zhang2025scizor}. These are important failure modes, but instruction-conditioned learning introduces another one that is behaviorally subtle and less explored: \emph{Instruction--Trajectory Mismatch} (ITM). We use ITM to describe episodes where the demonstrated behavior is coherent, but the paired natural-language instruction is semantically wrong (for example, labeling a \textit{pick} as a \textit{place}, or \textit{close the oven} as \textit{open the oven}). Such episodes can be harmful because they often look plausible in video and action space, pass standard quality filters, but can corrupt the mapping between language and behavior.

\begin{figure}[t]
    \includegraphics[trim={0.7cm 0.7cm 0.7cm 0.7cm},clip,width=1.0\linewidth]{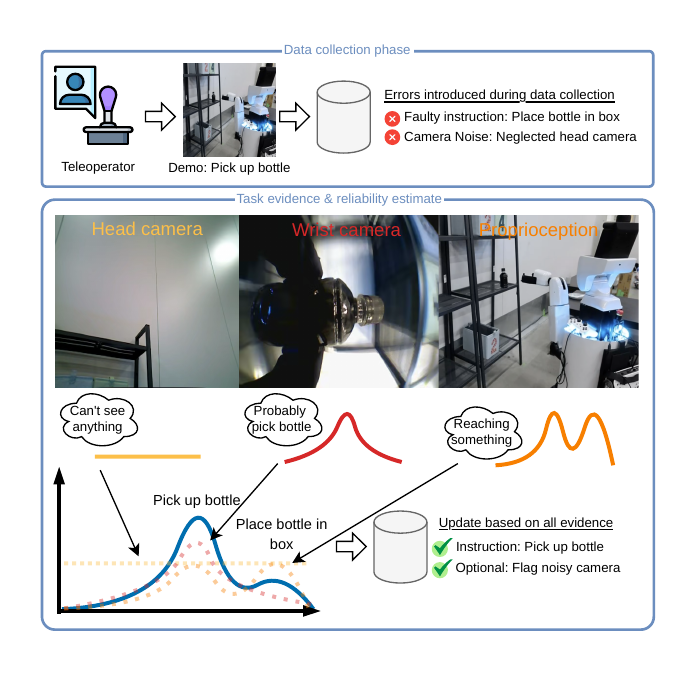}
    \caption{Example ITM auditing case. The provided instruction is ``Place bottle in box,'' while the trajectory is more consistent with ``Pick up bottle.'' MMPF estimates a task distribution and reliability score for each modality, then combines them using a reliability-weighted product of experts.}
    \label{fig:intro}
\end{figure}

Robot demonstrations are inherently \emph{multimodal}. Multiple camera views, proprioception, and potentially other sensors all provide partial evidence about the task. At the same time, robot data are collected in the real world, where any one modality can become unreliable at times because of occlusion, poor viewpoint, sensor corruption, or task-dependent ambiguity. A head camera may miss the relevant object, a wrist camera may be occluded, and proprioception may be weakly informative for tasks distinguished mainly by object state. Effective auditing therefore should not rely on a single view and it should aggregate across modalities to triangulate potential ITMs.

While recent work uses VLMs/LLMs to generate, relabel, or refine instructions for robot trajectories \cite{xiao2022robotic, kang2024clip, zhang2024sprint, glossop2025cast}, explicit \emph{post-hoc auditing} of instruction--trajectory mismatches in an already-labeled robot dataset remains under-explored, especially when considering multimodality with potential disagreement or unreliable observations.

In this work, we study \emph{offline task-instruction auditing}: given a robot dataset in which most labels are correct, identify episodes whose provided instruction does not match the demonstrated behavior. Our key hypothesis is that correctly labeled episodes lie in regions where multiple modalities provide consistent evidence for the provided task, whereas ITMs exhibit multimodal evidence that disagrees with the instruction. We propose \emph{Multimodal Probabilistic Fusion} (MMPF), a classifier-free auditing framework that combines local neighborhood agreement, global prototype similarity, and entropy-weighted multimodal fusion. The contribution of MMPF lies in integrating these established components into a multimodal ITM auditing framework, together with the associated problem formulation and evaluation. Fig.~\ref{fig:intro} illustrates the ITM auditing setting along with a conceptual view of our framework.

Our contributions are:
\begin{enumerate}
    \item We formalize ITM auditing for multimodal robot demonstration datasets.
    \item We introduce MMPF, a classifier-free framework that estimates task-label posteriors from embedding geometry and fuses modalities using reliability-weighted product-of-experts.
    \item We show that MMPF achieves the best overall ranking performance and correction accuracy across LIBERO and real-robot benchmarks.
    \item We show that auditing improves downstream policy learning most when language is needed for disambiguation, and that counterfactual evaluation reveals improved instruction grounding.
\end{enumerate}

\section{Related work}

\subsection{Robotics data curation}
A growing body of work studies data curation for imitation and language-conditioned robot learning. Some methods use task- or domain-specific quality heuristics, such as motion consistency or uncertainty measures~\cite{sakr2024consistency,valle2025evaluating}. Others estimate data utility directly through mutual information, transition-level progress, influence functions, datamodels, or learned success classifiers~\cite{hejna2025robot,zhang2025scizor,agia2025cupid,dass2025datamil,chenarss25}. Data-mixture methods such as ReMix~\cite{hejna2024re} optimize the composition of heterogeneous robot datasets. These approaches mainly target trajectory quality, redundancy, or utility and typically only investigate observations assuming task instructions are correct. In contrast, we target behaviorally valid demonstrations paired with semantically incorrect instructions.

\subsection{Task instruction labeling in robotics}
A complementary line of work uses pretrained vision-language models to synthesize task instructions for robot trajectories~\cite{xiao2022robotic,kang2024clip,zhang2024sprint,blank2024scaling,glossop2025cast}. These methods aim to create or enrich language supervision. This makes them valuable for scaling annotation, but less directly suited to the setting where a mostly correct human-designed dataset should be audited for semantic mismatches. Our goal is to preserve the human labor by auditing already labeled datasets, flag likely ITMs, and propose corrected task labels. Moreover, MMPF exploits the multimodality of robot data by treating each sensor stream as a separate expert with episode-dependent reliability.

\subsection{Label error detection}
Label-error detection has been studied through classifier-confidence methods such as Confident Learning~\cite{northcutt2021confident}, nearest-neighbor methods such as Deep kNN~\cite{bahri2020deep}, and multimodal label-error detection~\cite{zhang2024lemon}. Caption-error detection methods such as TRACED~\cite{afriat2025seeing} are also related, but target image-caption data using VLM-guided editing trajectories. Unlike these settings, robot ITM auditing naturally involves multiple sensors, intermittently uninformative observations, and reasoning over trajectories. MMPF performs post-hoc auditing directly from trajectory embeddings, without training a target classifier or running VLM inference.

\section{Method}\label{sec:method}
\begin{figure*}
    \centering
    \includegraphics[width=1.0\linewidth,trim={0.5cm 0.5cm 0.5cm 0cm},clip]{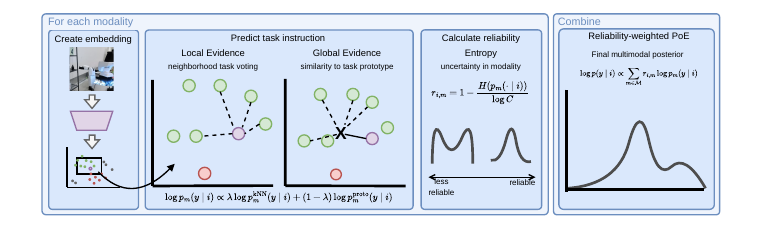}
    \caption{MMPF auditing framework. Each modality estimates a task-label distribution by combining local $k$-NN neighborhood agreement with global prototype similarity. Predictive entropy provides per-modality reliability weights for product-of-experts fusion; disagreement with the provided instruction indicates an ITM.}
    \label{fig:framework}
\end{figure*}

Given a dataset of $N$ robot episodes, each episode $i$ contains multimodal observations $x_{i,m}$ for modalities $m \in \mathcal{M}$ and a potentially noisy task label $\tilde{y}_i \in \{1,\dots,C\}$. A modality-specific encoder $f_m$ maps each trajectory to an embedding $z_{i,m} \in \mathbb{R}^{D_m}$. MMPF estimates a posterior $p(y \mid i)$ for the true task label by combining unimodal task evidence with an episode-specific reliability score $r_{i,m}$. The current formulation assumes a finite set of canonical task labels, with multiple episodes available for each task.

After acquiring the initial embeddings, MMPF has three steps as shown in Fig.~\ref{fig:framework}. Each modality first estimates task evidence from local neighborhoods and global prototypes with the evidence being combined into a final posterior. Then predictive entropy converts each posterior into a reliability score. Finally, a reliability-weighted product of experts fuses all modalities into a final multimodal posterior.

\subsection{Unimodal Evidence Estimation}
For each episode and modality, we build a per-modality evidence distribution $p_m(y\mid i)$ over task labels. This is performed by combining (a) local neighborhood agreement and (b) global prototype similarity.

\paragraph{Local evidence}
Let $\mathcal{N}_K(i,m)$ be the indices of the $K$ nearest neighbors of $z_{i,m}$ among all episodes in modality $m$,
under a distance $d^{(m)}(\cdot,\cdot)$ (e.g., Euclidean on $\ell_2$-normalized embeddings). The queried episode itself is excluded from the set.
We weight neighbors by an RBF kernel,
\begin{equation}
w^{(m)}_{ij}=\exp\!\left(-\frac{d^{(m)}(z_{i,m},z_{j,m})}{\sigma_m}\right), \qquad j\in\mathcal{N}_K(i,m),
\label{eq:knn_weights}
\end{equation}
where $\sigma_m > 0$ is a per-modality bandwidth parameter set to the mean pairwise distance in our experiments,
and convert the weighted label histogram into a distribution to predict the task label $y$:
\begin{equation}
p^{\mathrm{kNN}}_m(y\mid i)
=
\frac{\sum_{j\in\mathcal{N}_K(i,m)} w^{(m)}_{ij}\,\mathbf{1}[\tilde y_j=y]}{\sum_{y'=1}^{C}\sum_{j\in\mathcal{N}_K(i,m)} w^{(m)}_{ij}\,\mathbf{1}[\tilde y_j=y']}.
\label{eq:knn_dist}
\end{equation}

\paragraph{Global evidence}
Local neighborhoods can be noisy when data are sparse or when embeddings cluster by nuisance factors. To capture the global manifold structure, we compute a prototype $\mu_{c,m}$ for each class c (the mean embedding of all instances labeled $c$ in modality $m$). The global distribution is derived from the distance to these prototypes:
\begin{equation}
    p^{\text{proto}}_m(y \mid i) \propto \exp\left(-\frac{\|z_{i,m} - \mu_{y,m}\|}{\tau_m}\right)
    \label{eq:proto_prob}
\end{equation}
where $\tau_m > 0$ is a per-modality temperature parameter set to the mean prototype distance in our experiments.

\paragraph{Local and global combination}
We synthesize these two signals using a geometric mixture. This enforces a consensus constraint: a high probability requires support from both the local neighborhood and the global cluster center. We can control the trade-off with $\lambda\in[0,1]$:
\begin{equation}
\begin{aligned}
\log p_m(y \mid i) \propto {}&
\lambda \log\!\left(p^{\text{kNN}}_m(y \mid i)+\epsilon\right) \\
&+ (1-\lambda)\log\!\left(p^{\text{proto}}_m(y \mid i)+\epsilon\right).
\end{aligned}
\label{eq:pm}
\end{equation}

We renormalize after the geometric mixture. The small $\epsilon$ prevents zero-count kNN classes from producing undefined log probabilities. This yields the unimodal evidence distribution $p_m(y \mid i)$, which serves as the task label prediction for modality $m$.

\subsection{Modality reliability}
In teleoperated robotics, individual modalities may provide weak or ambiguous evidence independently of the task label (e.g., camera occlusion, sensor noise, or visually similar behaviors across tasks). We therefore compute a reliability score $r_{i,m} \in [0,1]$ to gate the contribution of each modality based on its predictive confidence. We measure confidence using the normalized entropy of the distribution:
\begin{equation}
    r_{i,m} = 1 - \frac{H(p_m(\cdot \mid i))}{\log C}.
\label{eq:conf_entr}
\end{equation}
Thus, sharp unimodal posteriors receive higher weight, while diffuse posteriors contribute less to fusion.

\subsection{Reliability-weighted multimodal fusion}
We fuse unimodal distributions with a reliability-weighted PoE, which emphasizes agreement across experts more strongly than a mixture-of-experts~\cite{hinton2002training}. Inspired by generalized PoE formulations~\cite{joshi2022gpoe,cao2014gpoe}, we use entropy-derived reliability weights:
\begin{equation}
    \log p(y \mid i) \propto \sum_{m \in \mathcal{M}} r_{i,m} \log p_m(y \mid i)
\label{eq:log_p}
\end{equation}

The final task prediction is then represented as:
\[\hat y_i = \arg\max_{y \in \{1,\dots,C\}} p(y \mid i)\]
In addition to task auditing, episodes with globally low per-modality reliability (where any or all $r_{i,m}$ are near zero) can additionally be flagged as \emph{noisy} regardless of the argmax prediction, indicating that there may be inherent problems in the observations. These could potentially be diagnosed manually or automatically filtered.

\section{Experiments}
We evaluate our method as an \emph{offline task-instruction auditing} tool for robot demonstration datasets. Our experiments target three questions:

\begin{enumerate}
    \item \textbf{Detection and correction:} Can the method identify and relabel instruction--trajectory mismatches more accurately than prior noisy-label baselines?
    \item \textbf{Robustness to modality corruption:} Does reliability-weighted multimodal fusion remain effective when a modality is degraded?
    \item \textbf{Downstream semantics:} When does better instruction auditing translate into better policy behavior, and when are the benefits better revealed by grounding-oriented evaluation than by standard clean-environment success alone?
\end{enumerate}

\subsection{Simulated setup}

\paragraph{Datasets}
We use LIBERO \cite{liu2023libero} and evaluate on four standard suites: LIBERO-10, LIBERO-Goal, LIBERO-Object, and LIBERO-Spatial. Each trajectory provides two camera views (head and wrist) and robot proprioception, which we treat as separate modalities.

To better isolate instruction following, we additionally evaluate on a counterfactual variant of LIBERO-Object, denoted \textbf{CF-Object}. This benchmark follows the evaluation design proposed in~\cite{fang2026visionoverrides}; because the original environment is not yet publicly available, we provide an independent re-implementation. The scene configuration is kept fixed but the task instruction is reassigned to a different object/task identity (e.g., the scene corresponds to \textit{pick up the alphabet soup}, while the instruction is changed to \textit{pick up the milk}). This breaks spurious correlations between scene layout and task identity, and therefore probes whether the policy truly relies on the language instruction rather than solving the task from visual context alone.

\paragraph{Hyperparameters}
All experiments use a fixed set of hyperparameters: $K{=}30$ nearest neighbors, RBF bandwidth $\sigma_m$ and temperature $\tau_m$ set to the mean pairwise distance for each modality, and mixing weight $\lambda{=}0.5$, which gives equal weight to local and global evidence and is not tuned per dataset. Trajectory embeddings are computed with Cosmos-Embed1~\cite{nvidia2024cosmosembed}, applied uniformly across all visual modalities. For robot states, because no comparable general-purpose
pretrained robot state encoder is available, we train a lightweight trajectory encoder and use its features as the embedding.
At each timestep, the encoder concatenates all normalized robot-state and action variables available for the corresponding platform. It processes the resulting sequence using a four-layer Transformer (width 256, four heads, feed-forward width 1024, dropout 0.1) with learned-query attention pooling. It is trained transductively on all trajectories in each audited dataset using masked state-delta reconstruction, masked action reconstruction, inverse dynamics, and forward-versus-reversed trajectory classification, with respective loss weights $1,1,0.5,0.5$. The resulting encoder is frozen before auditing. Task instructions and task labels are neither inputs nor loss targets; therefore, neither clean nor corrupted label information can leak into the encoder objective.

\paragraph{Noise Injection}
For simulations, we inject 30\% synthetic instruction-label noise into otherwise clean demonstrations. We consider two noise settings:

(i) \textbf{instance-dependent noise}. We make label flips more likely for trajectories near task boundaries. For each episode, we compute its nearest neighbors in trajectory-embedding space and identify episodes whose closest neighbors include demonstrations from different tasks. We then sample corrupted episodes from these boundary-near regions and replace their label with a confusable task label, chosen from the nearest different-task neighbor.

(ii) \textbf{clustered noise}. To simulate systematic annotation errors, we cluster demonstrations within each task into five clusters in embedding space. We then select one cluster per task and flip a fixed fraction of its labels to another task. Both synthetic corruption settings are constructed using the Cosmos-Embed1 video trajectory embeddings shared by all methods.

\paragraph{Baselines}
We compare MMPF against several noisy-label detection baselines:
\begin{itemize}
    \item \textbf{No Filtering:} train directly on the corrupted dataset.
    \item \textbf{Confident Learning (CL) \cite{northcutt2021confident}:} a widely used label-error detection method based on out-of-sample predicted probabilities from a supervised classifier.
    \item \textbf{Retrieval:} a retrieval-style baseline that scores instruction--trajectory consistency using vision-language similarity; our implementation adapts the DIAL instruction-augmentation approach~\cite{xiao2022robotic} to the auditing setting, applying CLIP-style scoring to the same pretrained trajectory embeddings used by MMPF.
    \item \textbf{LEMoN \cite{zhang2024lemon}:} a label-error detection method combining nearest-neighbor evidence in representation space and label-embedding space.
\end{itemize}
For a fair comparison, all baselines---including CL and LEMoN---are adapted to use the same Cosmos-Embed1~\cite{nvidia2024cosmosembed} trajectory and instruction embeddings used by MMPF, rather than on raw image features. For score-based baselines, decision thresholds are selected at the known synthetic noise prevalence. The Retrieval baseline additionally uses ridge regression to align the video and text embedding spaces. For the simulation, all end-policies are trained from scratch using smolvla~\cite{shukor2025smolvla} with three independent training seeds.

\paragraph{Evaluation metrics}
We report both threshold-free and decision-level auditing metrics. AUROC and AUPRC evaluate how well each method ranks mislabeled episodes above clean ones. Precision, recall, and F1 evaluate the actual filtering/relabeling decision. For MMPF, we report this at the method's native posterior-based decision rule. For score-based baselines, AUROC/AUPRC are computed directly from the continuous suspicion score, while F1 is reported at a prevalence-matched operating point under the synthetic 30\% noise setting. We additionally report \emph{correction accuracy}, defined among detected true ITMs: among episodes correctly flagged as mismatched, it measures the fraction for which the proposed replacement label matches the ground-truth task. Formally, $\mathrm{CorrAcc}=|\{i\in\mathcal{I}_{\mathrm{TP}}:\hat y_i=y_i\}|/|\mathcal{I}_{\mathrm{TP}}|$, where $\mathcal{I}_{\mathrm{TP}}$ denotes the set of correctly detected ITMs.

\subsection{Real-robot setup}
We evaluate on two datasets in real-robot settings shown in Fig.~\ref{fig:task_overview}. The first dataset, denoted \textit{Table}, decomposes similar tabletop behavior into primitive actions: focus, navigation, picking, and placing. It includes two objects, a paper and a mug where the tasks are to focus on, navigate to, pick up, and place the target object. In total it has 1260 demonstrations split across all primitive actions.

The second dataset, denoted \textit{Bottle-mug}, contains two families of two-stage
manipulation tasks in a shared tabletop/shelf setting: (i) picking up a bottle from the shelf and placing it into a target box, and (ii) picking up a mug and placing it at the missing corner of a rectangle on the table. Each episode includes head-camera observations, wrist-camera observations, and proprioception. This dataset has an additional 10\% observation noise where the teleoperator neglects the head camera making the scene partially observable. In total, the dataset includes 2980 demonstrations.

For the real-robot experiments, all policies are trained from scratch using pi0.5~\cite{intelligence2025pi_}. In addition, we generate 30\% uniform task instruction noise in the real-robot datasets. We use 10 evaluation rollouts per task for each real-robot policy. For the Table results, each stage reports the pooled success rate across both objects.

\subsection{Filtering instruction--trajectory mismatches}\label{sec:results}

\begin{table*}[t]
\centering
\caption{Instruction-auditing results on LIBERO benchmarks under instance-dependent and clustered label noise. Each cell reports \textbf{F1 / AUROC / AUPRC} (\%). Best is bold and second best is underlined for each metric within a row.}
\scriptsize
\setlength{\tabcolsep}{3.5pt}
\renewcommand{\arraystretch}{0.97}
\begin{tabular}{llcccc}
\toprule
Suite & Noise & MMPF (ours) & Conf. Learn. & Retrieval & LEMoN \\
\midrule
\multirow{2}{*}{LIBERO-10}
& Inst.  & \second{88.9} / \best{99.0} / \best{98.0} & 82.1 / 91.3 / 82.4 & \best{89.0} / 88.2 / 77.9 & 74.6 / \second{93.6} / \second{86.0} \\
& Clust. & \best{99.0} / \best{100.0} / \best{100.0} & \second{94.0} / 99.4 / 97.6 & 91.4 / 99.5 / 97.2 & 63.5 / \second{99.6} / \second{97.8} \\
\midrule
\multirow{2}{*}{LIBERO-Spatial}
& Inst.  & \best{99.6} / \best{99.8} / \best{99.4} & 64.8 / 74.6 / 61.9 & \second{79.9} / 78.3 / 61.2 & 60.8 / \second{83.1} / \second{71.8} \\
& Clust. & \best{95.5} / \best{99.9} / \best{99.6} & 62.7 / 79.9 / 55.5 & \second{68.3} / 74.1 / 49.0 & 60.3 / \second{80.1} / \second{60.7} \\
\midrule
\multirow{2}{*}{LIBERO-Object}
& Inst.  & \best{100.0} / \best{99.8} / \best{99.6} & 93.0 / \second{99.4} / \second{98.6} & \second{98.9} / 97.1 / 94.0 & 88.2 / 97.9 / 95.6 \\
& Clust. & \best{98.6} / \best{100.0} / \best{100.0} & 69.7 / 93.2 / 80.8 & \second{97.4} / 98.0 / 90.0 & 72.3 / \second{98.8} / \second{93.7} \\
\midrule
\multirow{2}{*}{LIBERO-Goal}
& Inst.  & \best{91.9} / \best{99.2} / \best{98.0} & 52.2 / 77.7 / \second{60.3} & \second{71.6} / 75.1 / 56.3 & 53.7 / \second{78.3} / 58.7 \\
& Clust. & \best{90.2} / \best{99.0} / \best{96.9} & 55.6 / 76.9 / 55.6 & \second{68.5} / 78.7 / 61.8 & 55.7 / \second{79.9} / \second{66.5} \\
\bottomrule
\end{tabular}
\label{tab:filter_main}
\end{table*}

\begin{table}[t]
\centering
\caption{Real-robot instruction-auditing results across the two robot datasets.}
\footnotesize
\setlength{\tabcolsep}{3pt}
\renewcommand{\arraystretch}{0.97}
\begin{tabular*}{\columnwidth}{@{\extracolsep{\fill}}lcccc@{}}
\toprule
\multicolumn{5}{c}{Decision-level metrics (\%)} \\
\midrule
Method & \multicolumn{2}{c}{Bottle-mug} & \multicolumn{2}{c}{Table} \\
\cmidrule(lr){2-3}\cmidrule(lr){4-5}
& F1 & Corr. & F1 & Corr. \\
\midrule
CL   & \second{66.8} & \second{67.2} & 78.3 & \second{85.6} \\
Ret. & 54.9 & 49.6 & \second{79.9} & 73.3 \\
LEMoN  & 60.9 & 35.4 & 77.5 & 73.4 \\
MMPF (ours) & \best{91.8} & \best{97.9} & \best{94.3} & \best{94.9} \\
\midrule
\multicolumn{5}{c}{Ranking metrics (\%)} \\
\midrule
Method & \multicolumn{2}{c}{Bottle-mug} & \multicolumn{2}{c}{Table} \\
\cmidrule(lr){2-3}\cmidrule(lr){4-5}
& AUROC & AUPRC & AUROC & AUPRC \\
\midrule
CL   & \second{99.6} & \second{99.4} & \second{98.5} & \second{96.9} \\
Ret. & 83.6 & 60.4 & 87.8 & 76.1 \\
LEMoN  & 83.9 & 66.0 & 92.3 & 86.3 \\
MMPF (ours) & \best{99.9} & \best{99.8} & \best{99.6} & \best{99.3} \\
\bottomrule
\end{tabular*}
\label{tab:filter_robotics}
\end{table}

We first evaluate MMPF as an offline auditor for identifying instruction trajectory mismatches under \emph{instance-dependent} and \emph{clustered} synthetic label noise. We report AUROC and AUPRC as threshold-free ranking metrics, together with F1 at each method. We further provide correction accuracy summarized for LIBERO and reported for the real-robot benchmarks. Table~\ref{tab:filter_main} summarizes the main LIBERO results, and Table~\ref{tab:filter_robotics} reports the real-robot results.

Across the LIBERO suites, MMPF provides the strongest \emph{overall} auditing signal. Its clearest advantage is on threshold-free ranking, where MMPF consistently achieves the best AUROC/AUPRC, indicating that its multimodal posterior orders mismatched episodes above clean ones more reliably than the baselines. This advantage is especially important on the harder semantic settings. For example, on LIBERO-10 with instance-dependent noise, MMPF reaches 99.0 AUROC and 98.0 AUPRC, versus 88.2 and 77.9 for retrieval, even though retrieval is narrowly higher for F1 (89.0 vs.\ 88.9). On LIBERO-Spatial, LIBERO-Goal, and LIBERO-Object, MMPF is strongest on both ranking and practical filtering metrics, including near 100 F1 on several settings.

Correction quality is another major strength. Across all eight LIBERO settings in Table~\ref{tab:filter_main}, MMPF attains 100.0\% correction accuracy among detected true ITMs. The baselines are often competitive but less consistent: retrieval ranges from 76.2\% to 98.1\%, Confident Learning from 72.4\% to 100.0\%, and LEMoN from 72.5\% to 100.0\%. In other words, MMPF does not merely rank suspicious episodes well; when it proposes a replacement label, that proposal is also highly reliable. The high correction accuracy on LIBERO is aided by its controlled task structure, where task identities are relatively well separated and the synthetic mismatches follow the trajectory-embedding geometry. The real-robot experiments provide a more challenging complement, using uniform label noise across behaviorally similar tasks, while still showing strong
correction performance.

The same pattern transfers beyond simulation. Across both real-robot datasets, MMPF achieves the strongest overall auditing performance. On the bottle-mug dataset, MMPF reaches 91.8 F1, 99.9 AUROC, and 99.8 AUPRC, while also obtaining the highest correction accuracy at 97.9\%. On the table dataset, MMPF again performs best, reaching 94.3 F1, 99.6 AUROC, 99.3 AUPRC, and 94.9 correction accuracy. This suggests that MMPF provides both reliable ranking of suspicious demonstrations and accurate correction proposals across real-robot settings.

These results support our central hypothesis: for ITM auditing in robotics, it is beneficial to combine local neighborhood agreement and global prototype structure across modalities.

\subsection{Downstream policy learning after auditing}
We next examine how instruction-label quality translates into better instruction-conditioned policy learning. Our belief is that after auditing, the dataset should better align language with demonstrated behavior. Whether this also improves standard clean-environment success depends on how strongly the benchmark requires language to disambiguate the task. Recent work shows that policies can sometimes recover the intended behavior from scene context alone, even when instruction grounding is imperfect~\cite{fang2026visionoverrides}. For each noisy training set, we apply each auditing method, filter out the flagged demonstrations, train a policy, and report clean-environment success rates in Table~\ref{tab:policy_main}.

The clearest gains appear on suites where preserving the mapping between language and behavior matters most. MMPF improves success substantially on LIBERO-Goal (77.5 vs.\ 58.7 for no filtering) and LIBERO-Spatial (76.7 vs.\ 70.3), indicating that removing instruction--trajectory mismatches can materially improve policy learning when task semantics are harder to infer from context alone. The retrieval baseline reaches 66.0 on LIBERO-Goal and 68.0 on LIBERO-Spatial---above no filtering on both---but remains well below MMPF, suggesting that CLIP-style instruction--trajectory scoring alone is not sufficient for the harder semantic mismatch cases.

At the same time, the gains are not uniform across all suites. On LIBERO-10, MMPF (51.2), retrieval (50.5), and the unfiltered baseline (51.0) perform similarly, while Confident Learning is marginally higher at 52.3. On LIBERO-Object, the unfiltered baseline and LEMoN both reach 86.2, slightly above MMPF at 83.2. Our interpretation is that some suites allow the policy to exploit regularities in scene layout and object configuration, reducing its dependence on the instruction at test time even when the training labels are semantically cleaner. We therefore perform a grounding-oriented analysis below on LIBERO-Object.

\begin{table*}[t]
\centering
\caption{Clean-evaluation policy success (\%). Each cell reports the mean and 95\% Student-$t$ confidence interval over three independent policy-training seeds. Intervals are clipped to $[0,100]$. Best point estimate is bold and second best is underlined.}
\footnotesize
\setlength{\tabcolsep}{4pt}
\renewcommand{\arraystretch}{0.96}
\begin{tabular*}{\textwidth}{@{\extracolsep{\fill}}lccccc@{}}
\toprule
Method & L10 & Goal & Spatial & Object & CF-Obj \\
\midrule
No filt. & 51.0 [41.6,60.4] & 58.7 [50.6,66.7] & \second{70.3} [58.3,82.4] & \best{86.2} [75.6,96.7] & 15.0 [6.0,24.0] \\
CL       & \best{52.3} [46.0,58.7] & 62.8 [60.2,65.4] & 66.2 [53.5,78.9] & \second{85.8} [80.8,90.9] & \second{30.8} [28.1,33.5] \\
Ret.     & 50.5 [43.6,57.4] & \second{66.0} [59.5,72.5] & 68.0 [58.1,77.9] & 85.0 [67.6,100.0] & 29.5 [24.1,34.9] \\
LEMoN    & 17.0 [9.4,24.6] & 45.3 [39.6,51.1] & 66.7 [60.5,72.8] & \best{86.2} [84.3,88.1] & 24.5 [22.0,27.0] \\
MMPF (ours) & \second{51.2} [39.9,62.4] & \best{77.5} [71.8,83.2] & \best{76.7} [62.7,90.6] & 83.2 [78.6,87.8] & \best{30.9} [25.8,36.0] \\
\bottomrule
\end{tabular*}
\label{tab:policy_main}
\end{table*}

\subsection{Grounding-oriented evaluation}
To test whether the downstream gains reflect better language grounding rather than only generic control improvements, we evaluate on CF-Object, which follows the evaluation paradigm of~\cite{fang2026visionoverrides} that proposes counterfactual benchmark construction to isolate genuine instruction dependence. A policy in this benchmark cannot simply rely on familiar visual context and must instead use the language instruction to select the correct object. Table~\ref{tab:grounding_main} reports both task success and object-level grounding metrics.

MMPF consistently reduces interaction with the biased object while maintaining or improving interaction with the instructed object. The more interesting comparison is with the no filtering baseline which had the best performance on LIBERO-Object. MMPF increases faithful grasp rate from 28.3\% to 41.4\% and reduces biased grasp rate from 40.3\% to 14.3\% over this baseline. This suggests that the main downstream benefit of auditing is not merely better low-level control, but a cleaner alignment between task language and demonstrated behavior.

\begin{table}[t]
\centering
\caption{Grounding-oriented evaluation on CF-Object. F-T/F-G denote faithful touch/grasp rates (higher is better); B-T/B-G denote biased touch/grasp rates (lower is better). Best is bold and second best is underlined.}
\scriptsize
\setlength{\tabcolsep}{3pt}
\renewcommand{\arraystretch}{0.96}
\begin{tabular}{lcccc}
\toprule
Method & F-T $\uparrow$ & F-G $\uparrow$ & B-T $\downarrow$ & B-G $\downarrow$ \\
\midrule
Base & 29.0 & 28.3 & 42.0 & 40.3 \\
CL   & 40.6 & 40.6 & 19.8 & 19.6 \\
Ret. & \second{41.8} & \second{41.2} & \second{18.5} & \second{18.2} \\
LEMoN  & 39.5 & 38.2 & 25.7 & 24.8 \\
MMPF (ours) & \best{42.2} & \best{41.4} & \best{14.4} & \best{14.3} \\
\bottomrule
\end{tabular}
\label{tab:grounding_main}
\end{table}

\subsection{Ablation study}

\begin{table}[t]
\centering
\caption{Ablation of MMPF. Full MMPF attains the best overall F1.}
\small
\setlength{\tabcolsep}{4pt}
\begin{tabular}{lccc}
\toprule
Variant & Prec $\uparrow$ & Rec $\uparrow$ & F1 $\uparrow$ \\
\midrule
Proprio only                & 97.5 & 84.2 & 90.4 \\
Visual only (all cameras)   & 96.6 & 85.5 & 90.7 \\
Local only (kNN)            & \best{99.9} & 81.1 & 89.5 \\
Global only (prototype)     & 33.3 & \best{100.0} & 49.9 \\
No reliability weighting    & 99.7 & 81.0 & 89.4 \\
Multimodal kNN only         & 99.7 & 80.9 & 89.3 \\
MMPF (full)                 & \best{99.9} & 84.9 & \best{91.8} \\
\bottomrule
\end{tabular}
\label{tab:ablation}
\end{table}

To assess the contribution of each design choice, we compare MMPF against ablated variants on the Bottle-mug robot benchmark (Table~\ref{tab:ablation}), which best exposes differences in modality informativeness and reliability. Full MMPF achieves the best overall balance at 99.9\% precision, 84.9\% recall, and 91.8 F1. Restricting the method to a single modality group lowers F1 to 90.4 with proprioception only and 90.7 with visual observations only, showing that the modalities provide complementary evidence. Likewise, removing either evidence source hurts performance: Local only (kNN) falls to 89.5 F1, while Multimodal kNN only reaches 89.3, indicating that the local and global signals are both useful when combined properly.

Removing reliability weighting reduces F1 from 91.8 to 89.4, mainly by lowering recall while keeping precision nearly unchanged. This supports the role of reliability weighting as a per-episode gate that suppresses uncertain modalities rather than simply sharpening predictions globally. The most extreme failure is the prototype-only variant, which attains 100.0\% recall but only 33.3\% precision, collapsing to 49.9 F1. However, by making use of the local neighborhood agreement, this balances out the coarseness of the global evidence improving the precision.

\begin{table}[t]
\centering
\caption{Effect of reliability thresholding on noisy-episode detection.
We vary the threshold to mark noisy observations for the Bottle-mug setting.}
\small
\setlength{\tabcolsep}{5pt}
\begin{tabular}{lcccc}
\toprule
\textbf{Min Rel.} & \textbf{Prec} $\uparrow$ & \textbf{Rec} $\uparrow$ & \textbf{Rec Obs} $\uparrow$ & \textbf{Rec Label} $\uparrow$ \\
\midrule
0.0 & 99.9\% & 81.3\%  & 13.3\% & 99.2\% \\
0.1 & 99.9\% & 84.9\%   & 30.0\% & 99.2\% \\
0.2 & 97.3\% & 95.1\%   & 77.8\% & 99.2\% \\
\bottomrule
\end{tabular}
\label{tab:reliability_threshold_ablation}
\end{table}

Table~\ref{tab:reliability_threshold_ablation} shows that the same reliability signal used inside MMPF can also be thresholded to detect sensor-corrupted episodes. The table shows the combined precision and recall when it comes to detecting label error or sensor-corruption. Raising the threshold from 0.0 to 0.2 increases overall recall from 81.3\% to 95.1\%, driven mainly by better recovery of episodes with sensor-corruption (13.3\% to 77.8\%). In the main results we use the fixed conservative threshold of 0.1; Table~\ref{tab:reliability_threshold_ablation} reports sensitivity to this operating point. This supplementary feature can be used to analyze episodes that might have one or more broken sensors.

\subsection{Robot experiment}
\begin{figure}[t]
    \centering
    \setlength{\tabcolsep}{1pt}
    \begin{tabular}{ccc}
        \includegraphics[width=0.31\columnwidth,trim={0.3cm 0.4cm 0.3cm 0.4cm},clip]{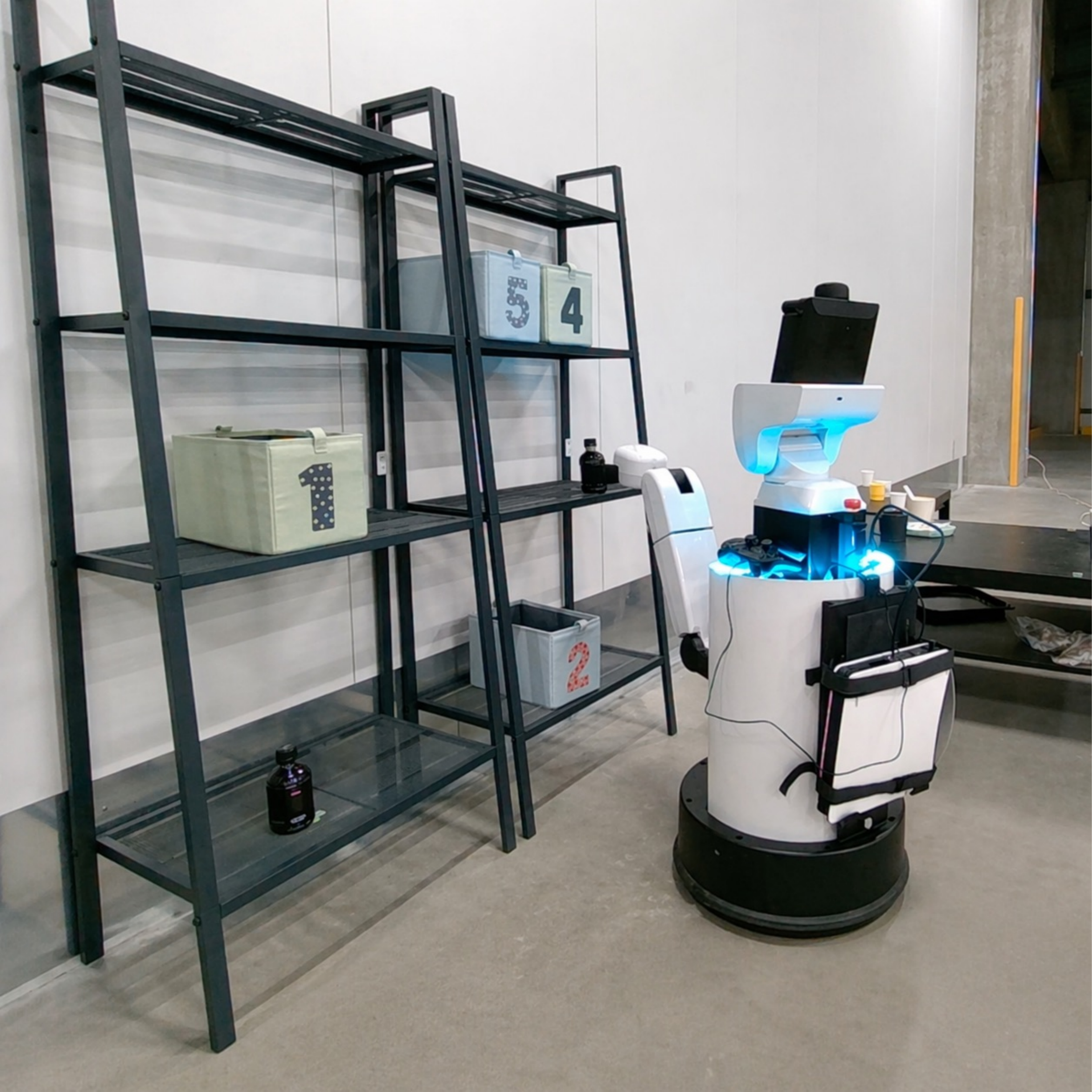} &
        \includegraphics[width=0.31\columnwidth,trim={0.3cm 0.4cm 0.3cm 0.4cm},clip]{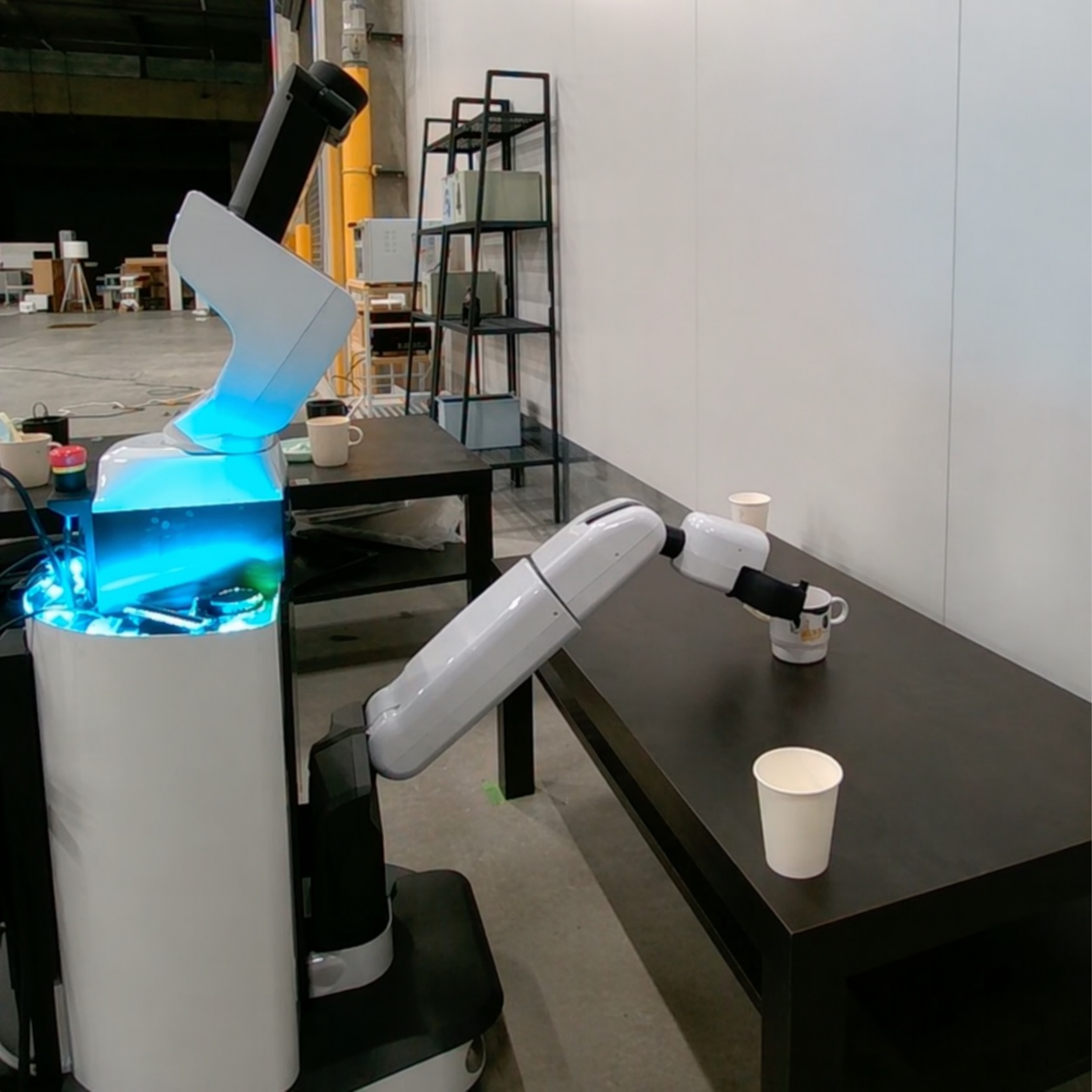} &
        \includegraphics[width=0.31\columnwidth,trim={0.3cm 0.4cm 0.3cm 0.4cm},clip]{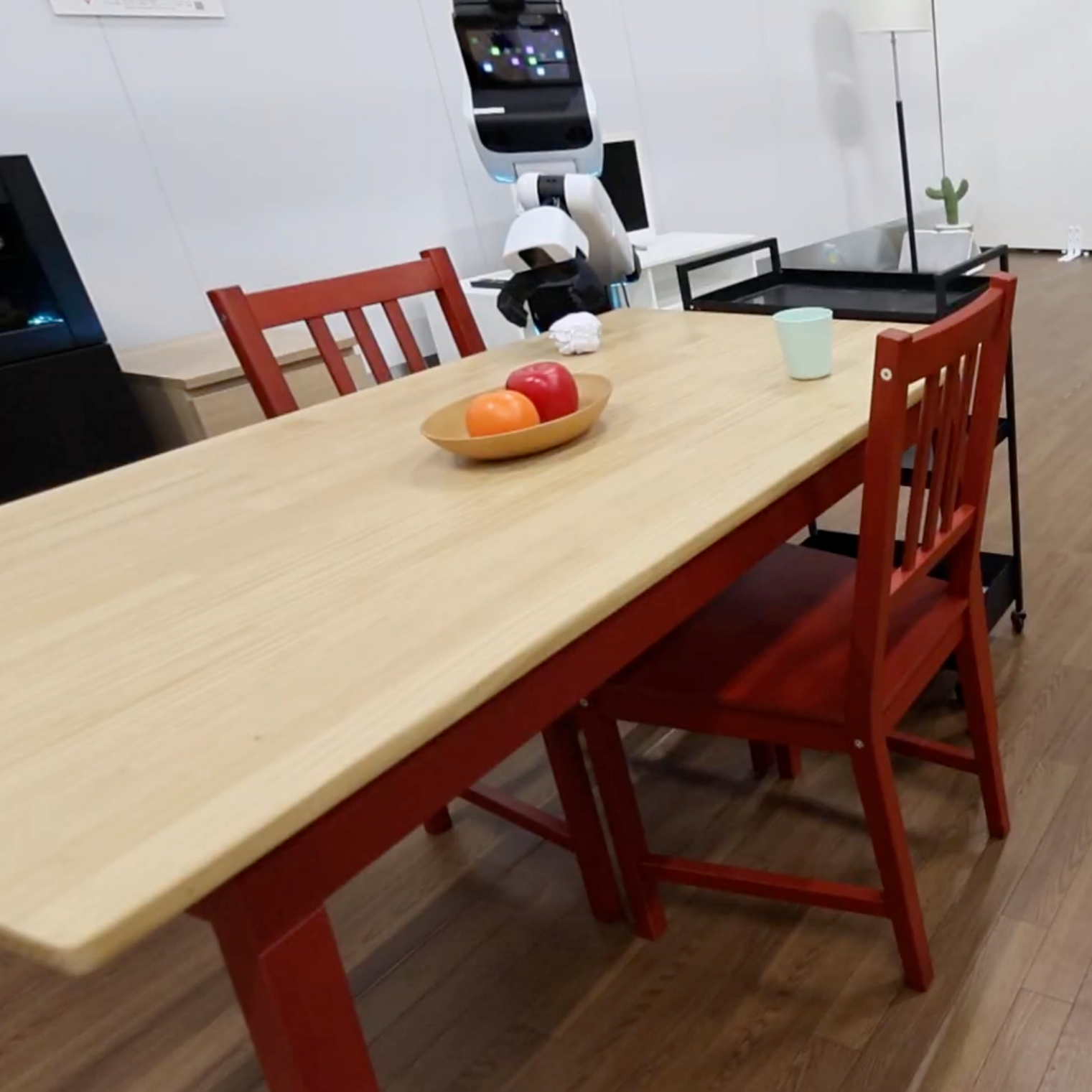} \\
        \scriptsize Bottle & \scriptsize Mug & \scriptsize Table
    \end{tabular}
    \caption{Real-robot environments used for downstream evaluation.}
    \label{fig:task_overview}

\end{figure}

We conduct two real-robot downstream evaluations. On the Table dataset, we evaluate conservative filtering, where demonstrations flagged as ITMs are removed from the training set. For both the Table dataset and Bottle-mug dataset we use correction-based curation where the demonstrations are kept but relabeled with the suggested task instruction.

\begin{table}[t]
\centering
\caption{Real-robot downstream policy success (\%). Each Table stage pools 20 rollouts (10 per object), while each Bottle-mug stage contains 10 rollouts. The ``All'' column reports pooled success with 95\% Wilson confidence intervals over 80 rollouts per Table condition and 40 per Bottle-mug condition.}
\label{tab:real_robot_policy}
\scriptsize
\setlength{\tabcolsep}{3pt}
\renewcommand{\arraystretch}{0.92}

\begin{tabular*}{\columnwidth}{@{\extracolsep{\fill}}lccccc@{}}
\toprule
\multicolumn{6}{c}{Table: filtering-based cleaning} \\
\midrule
Method & All & Focus & Nav. & Pick & Place \\
\midrule
Unfilt. & \resultci{73.8}{63.2,82.1} & 95.0 & 85.0 & 65.0 & 50.0 \\
CL      & \resultci{67.5}{56.6,76.8} & 85.0 & 80.0 & 45.0 & \best{60.0} \\
MMPF (ours) & \resultci{\best{78.8}}{68.6,86.3} & \best{100.0} & \best{100.0} & \best{75.0} & 40.0 \\
\end{tabular*}

\begin{tabular*}{\columnwidth}{@{\extracolsep{\fill}}lccccc@{}}
\toprule
\multicolumn{6}{c}{Table: correction-based cleaning} \\
\midrule
Method & All & Focus & Nav. & Pick & Place \\
\midrule
Unfilt. & \resultci{73.8}{63.2,82.1} & 95.0 & 85.0 & 65.0 & 50.0 \\
CL      & \resultci{85.0}{75.6,91.2} & \best{100.0} & \best{100.0} & \best{100.0} & 40.0 \\
MMPF (ours) & \resultci{\best{90.0}}{81.5,94.8} & \best{100.0} & \best{100.0} & 95.0 & \best{65.0} \\
\bottomrule
\end{tabular*}

\vspace{0.35em}

\begin{tabular*}{\columnwidth}{@{\extracolsep{\fill}}lccccc@{}}
\toprule
\multicolumn{6}{c}{Bottle-mug: correction-based cleaning} \\
\midrule
Method & All & Bottle pick & Bottle place & Mug pick & Mug place \\
\midrule
Unfilt. & \resultci{62.5}{47.0,75.8} & 90 & 70 & 70 & 20 \\
CL      & \resultci{60.0}{44.6,73.7} & \best{100} & 50 & 60 & \best{30} \\
MMPF (ours) & \resultci{\best{77.5}}{62.5,87.7} & \best{100} & \best{100} & \best{80} & \best{30} \\
\bottomrule
\end{tabular*}
\end{table}

Table~\ref{tab:real_robot_policy} reports the downstream policy success rates. Looking first at filtering-based curation in the Table environment, MMPF obtains the highest observed average success across all primitive tasks, with a point estimate of 78.8\% compared with 73.8\% for the unfiltered baseline and 67.5\% for CL. Correction-based curation yields a higher observed point estimate: when detected instruction errors are relabelled rather than removed, MMPF obtains an average success rate of 90.0\%, compared with 85.0\% for CL and 73.8\% for the unfiltered baseline. Both CL and MMPF reach near-ceiling performance on the easier focus, navigation, and pick stages, while MMPF has the highest observed overall average and placement success under correction-based cleaning.

The combined results reveal an important trade-off in small real-robot datasets. Conservative filtering can improve semantically grounded stages by removing demonstrations with unreliable instruction labels, as seen in the strong focus and navigation performance of MMPF even after data removal. However, filtering also reduces the amount of available demonstration data, which can hurt stages that depend more strongly on low-level control coverage. This is most visible in the placement task: MMPF obtains only 40.0\% placement success under filtering, but improves to 65.0\% when the same type of detected mismatches are corrected instead of discarded. Thus, even with imperfect correction accuracy, correction-based curation can be preferable when the underlying trajectory remains a valid demonstration.

The Bottle-mug dataset provides a more challenging correction-based setting. MMPF obtains the highest observed average success, reaching 77.5\% compared with 62.5\% for the unfiltered baseline and 60.0\% for CL. CL does not improve over the unfiltered baseline on average, which is consistent with its lower correction accuracy of 67.2\% compared with 97.9\% for MMPF. Overall, these results show that MMPF is useful not only for detecting instruction--trajectory mismatches, but also for producing accurate correction labels. This becomes especially important in small robot datasets, where preserving valid control trajectories can be as important as removing semantically incorrect supervision.

\section{Conclusion}

We presented an offline auditing framework for detecting Instruction--Trajectory Mismatches (ITM) in multimodal robot demonstration datasets. Our method Multimodal Probabilistic Fusion requires no clean-label supervision and operates directly on pretrained trajectory embeddings. By treating each modality as an independent expert and gating its contribution via an entropy-based reliability score, the framework remains effective even when individual modalities provide weak or ambiguous evidence.

Across synthetic and real-robot benchmarks, MMPF achieves the strongest \emph{overall} ITM auditing performance. It consistently provides the best threshold-free detection scores, while also delivering strong practical cleaning performance. Reliability weighting improves performance while the minimum-reliability threshold offers an additional mechanism for surfacing potentially corrupt episodes in real-robot data. Downstream, filtering improves policy success rates on suites where language is a necessary guide for action selection (LIBERO-Goal, LIBERO-Spatial, CF-Object). The CF-Object grounding evaluation further shows that one primary benefit of auditing is improved instruction--behavior alignment rather than generic control improvement. We further showed in real-world experiments that we can attain an overall better success rate across tasks using MMPF. We showed that by performing correction-based curation rather than filtering, we improve the trade-off between learning low-level control behavior while maintaining semantic understanding.

A limitation of MMPF is that it depends on the quality of the underlying trajectory embeddings. The post-embedding auditing stage is training-free, but it still requires representations in which task-relevant behavior is geometrically meaningful. Pretrained video and world-model embeddings are increasingly available, but less so for robot proprioception, where broadly reusable pretrained encoders are not yet standard. In our experiments, this required training a lightweight robot-state trajectory encoder. MMPF also assumes a finite set of canonical task instructions and performs auditing as discrete label correction. This assumption covers many existing robot instruction datasets and related relabeling pipelines, but does not yet address fully open-vocabulary task descriptions. Future work could support open-vocabulary descriptions by replacing hard task classes with soft neighborhoods in instruction-embedding space, allowing semantically equivalent paraphrases to share evidence. Performing this extension on an open-ended VLA dataset with paraphrases, scene diversity, and naturally occurring label noise remains future work. Another interesting direction is to utilize automatic label synthesising methods with our auditing framework to improve the integrity of generated instructions. Finally, there is no limit in how many modalities can be used so future work could investigate how adding more modalities like touch could improve auditing.

\bibliographystyle{IEEEtran}
\bibliography{citations}

\end{document}